\documentclass[10pt,twocolumn,letterpaper]{article}

\newcommand{\Tref}[1]{Table~\ref{#1}}

\newcommand{\fref}[1]{Fig.~\ref{#1}}

\newcommand{\sref}[1]{Sec.~\ref{#1}}

\usepackage{cvpr}
\usepackage{times}
\usepackage{epsfig}
\usepackage{graphicx}
\usepackage{amsmath}
\usepackage{amssymb}

\usepackage{enumitem}
\usepackage{xfrac}
\usepackage{ctable}
\usepackage{array}
\usepackage{caption}
\usepackage{multirow}
\usepackage{subfig}
\usepackage{tabularx}
\usepackage{capt-of}

\renewcommand{\paragraph}[1]{\vspace{1mm}\noindent\textbf{#1}}

\newcolumntype{P}[1]{>{\centering\arraybackslash}p{#1}}
\newcolumntype{M}[1]{>{\centering\arraybackslash}m{#1}}
\newcolumntype{?}[1]{!{\vrule width #1}}

\newlength{\Oldarrayrulewidth}

\usepackage[pagebackref=true,breaklinks=true,letterpaper=true,colorlinks,bookmarks=false]{hyperref}

\cvprfinalcopy % *** Uncomment this line for the final submission

\def\cvprPaperID{2338} % *** Enter the CVPR Paper ID here

\ifcvprfinal\fi
\begin{document}

%%%%%%%%% TITLE
\title{Fast User-Guided Video Object Segmentation \\ by Interaction-and-Propagation Networks}

\author{
Seoung Wug Oh\\Yonsei University \and Joon-Young Lee\\Adobe Research \and  Ning Xu\\Adobe Research \and  Seon Joo Kim\\Yonsei University
}

\maketitle
\thispagestyle{empty}

%%%%%%%%% ABSTRACT
\begin{abstract}
We present a deep learning method for the interactive video object segmentation. Our method is built upon two core operations, interaction and propagation, and each operation is conducted by Convolutional Neural Networks. The two networks are connected both internally and externally so that the networks are trained jointly and interact with each other to solve the complex video object segmentation problem. We propose a new multi-round training scheme for the interactive video object segmentation so that the networks can learn how to understand the user's intention and update incorrect estimations during the training. At the testing time, our method produces high-quality results and also runs fast enough to work with users interactively. 
We evaluated the proposed method quantitatively on the interactive track benchmark at the DAVIS Challenge 2018. We outperformed other competing methods by a significant margin in both the speed and the accuracy. We also demonstrate that our method works well with real user interactions. 
\end{abstract}

%%%%%%%%% BODY TEXT
\section{Introduction}
Video object segmentation is a task of separating a foreground object from a video sequence. It is an essential task in video editing with a wide range of applications from the consumer-level video editing to the professional TV and movie post-production.
This problem is often solved by either a fully-automatic approach (\ie unsupervised foreground object segmentation~\cite{tokmakov2017learning}) or a semi-supervised approach (\ie ground-truth object masks are given on few frames~\cite{caelles2017one, perazzi2017learning}).
However, both solutions have limitations in reflecting a user's intention or refining incorrect estimations.

%%%%%% figure %%%%%%%%%%%
\begin{figure}
\centering
\includegraphics[width=1.0\linewidth]{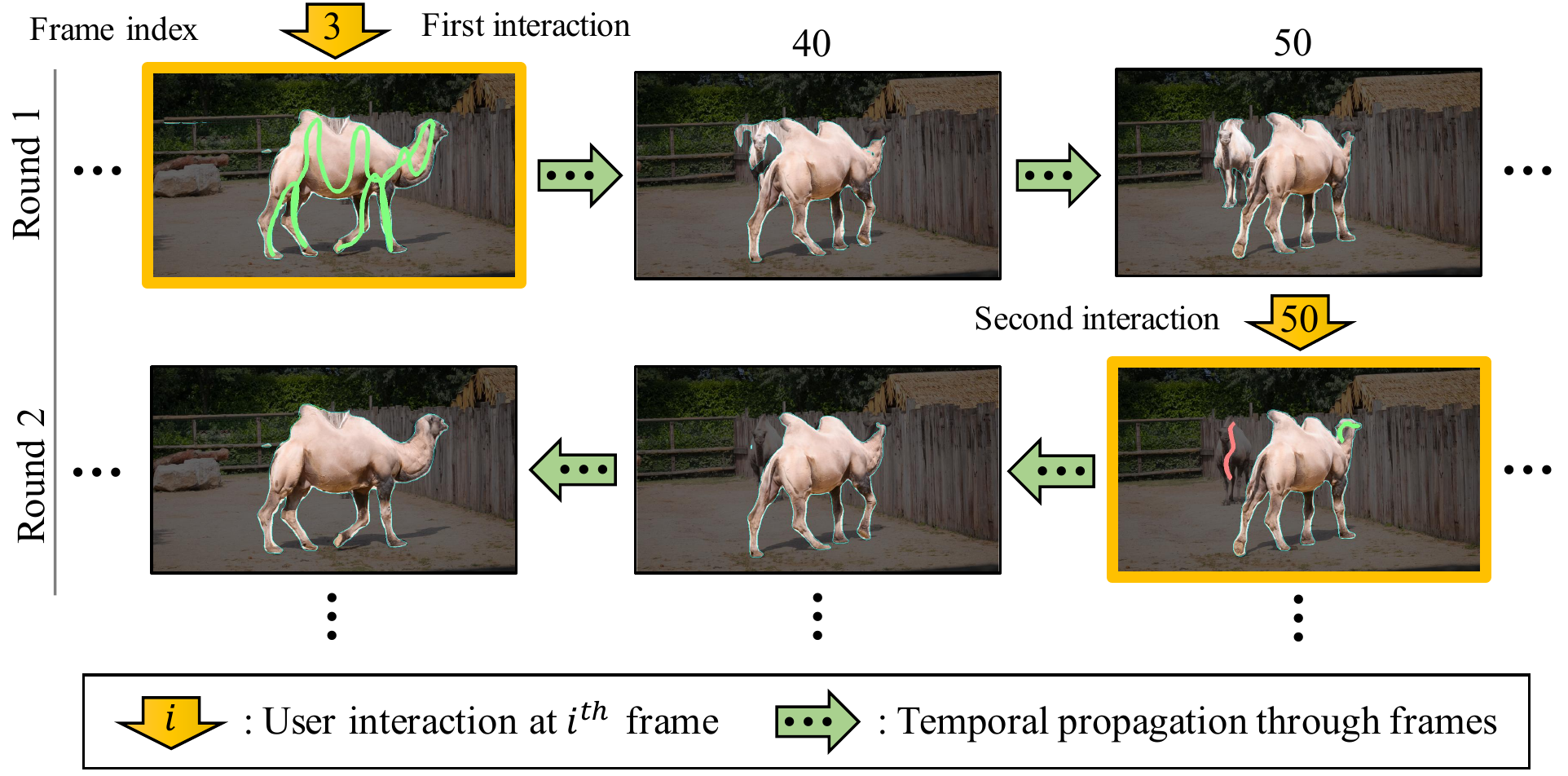}
\caption{We propose a method that can estimate object masks in a video by interacting with a user. 
The mask of a target object is generated using user annotations (\eg scribbles at frame 3), and the computed mask is propagated to compute the masks for the entire video. 
The user can repeatedly provide additional feedback (\eg scribbles on false positive and false negative at frame 50) to refine the segmentation masks.
Our method generates high-quality object masks with minimal user interactions and time budget.}
\label{Fig:teaser}
\end{figure}
%%%%%%%%%%%%%%%%%%%%%%%%%

Interactive video segmentation can potentially resolve this issue by allowing user intervention given in a user-friendly form such as scribbles~\cite{wang2005interactive, price2009livecut, bai2009video}.
However, existing interactive methods require a lot of user interactions to obtain results with acceptable quality for video editing applications.
In this paper, we aim to develop an interactive video object segmentation technique that can estimate accurate object masks in a video sequence with minimal user interactions.

%In practice, video cutout is laboriously done by skillful artists with significant manual interactions by using commercial tools like Adobe %After Effect. In addition, consumer video sharing (\eg Instagram, Facebook, Snapchat) is becoming very popular and there is a strong demand for %easy-to-use consumer-level tools for video editing.
%To address these issues, we aim to develop an interactive video object segmentation method that can estimate accurate object masks in a video %sequence with minimal user interactions.

Interactive video cutout methods usually follow the procedure of the rotoscoping~\cite{bratt2012rotoscoping, li2016roto++}, where a user sequentially processes a video frame-by-frame. 
In this scenario, the user verifies and updates the object mask with multiple interactions at every frame. 
%To reduce the user interaction, this process is done sequentially along the sequence frame order and an algorithm may initialize an object mask %by propagating the previous mask. 
This rotoscoping-style interaction requires a lot of effort and is more suitable for professional uses that require high-quality results. 
%a user decides to whether continue to the next frame or refine the current frame with additional interaction at every frame. 

Recently, Caelles~\etal~\cite{caelles20182018} introduced another workflow for the video object cutout that can minimize the user's effort. In this scenario, which we call as the \textit{round-based interaction}, the user provides annotations on a selected frame and an algorithm computes the segmentation maps for all video frames in a batch process.
To refine the results, the process of user annotation and segmentation map computations are repeatedd until the user is satisfied with the results. 
This round-based interaction is useful for consumer-level applications and rapid prototyping for professional usage, where the efficiency is the main concern. 
One can control the quality of the segmentation according to the time budget, as more rounds of interactions will provide more accurate results. 

In this paper, we present a deep learning based method for the interactive video object segmentation
tailored to the round-based interaction scenario (\fref{Fig:teaser}).
While several deep learning approaches for video object segmentation have been proposed~\cite{caelles2017one, perazzi2017learning}, they are usually too slow for the interactive scenario as they rely heavily on online learning.
Even with a fast video segmentation algorithm~\cite{oh2018fast}, designing a deep neural network (DNN) and its training mechanism for the interactive segmentation scenario remains as a challenge.

To solve this challenging problem, we propose the Interaction-and-Propagation Networks and an effective training method.
Our framework consists of two deep CNNs, each of which is dedicated to the core operations \textit{interaction} and \textit{propagation} respectively.
The interaction network takes the user annotation (\eg scribbles) to segment the foreground object. The propagation network transfers the object mask computed in the source frame to other neighboring frames.
These two networks are internally connected using our feature aggregation module and are also externally connected so that each of them takes the other's output as its input.

%We define two core operations, \textit{interaction} and \textit{propagation}, and construct two deep networks dedicated to each operation (see %\fref{Fig:networks}).
%The interaction network takes the user annotation (\eg scribbles) to segment the foreground object. The propagation network transfers the object %mask computed in the source frame to other neighboring frames.
%These two networks are internally connected using our feature aggregation module and are also externally connected so that each of them takes %the other's output as their input.

The two networks are trained jointly to adapt to each other, which reduces unstable behaviors between the two operations.
We also propose the concept of multi-round training, which is specifically designed to simulate a real testing scenario of the interactive video segmentation.
In this training strategy, a number of user feedback cycles and the response of networks form a single training iteration (see \fref{Fig:multi-round}). This new training scheme greatly improves the performance of our model. 

Our framework is quantitatively evaluated on the interactive track benchmark at the DAVIS Challenge 2018~\cite{caelles20182018} and achieves the state-of-the-art performance with a big gap compared to other competing methods~\cite{DAVIS2018-Interactive-1st}.
We also demonstrate the usefulness of our method with real interactive cutout use-cases.
We will release the source code that contains our trained model and the graphical user interface.

%%%%%% figure %%%%%%%%%%%
\begin{figure*}
\centering
\includegraphics[width=\linewidth]{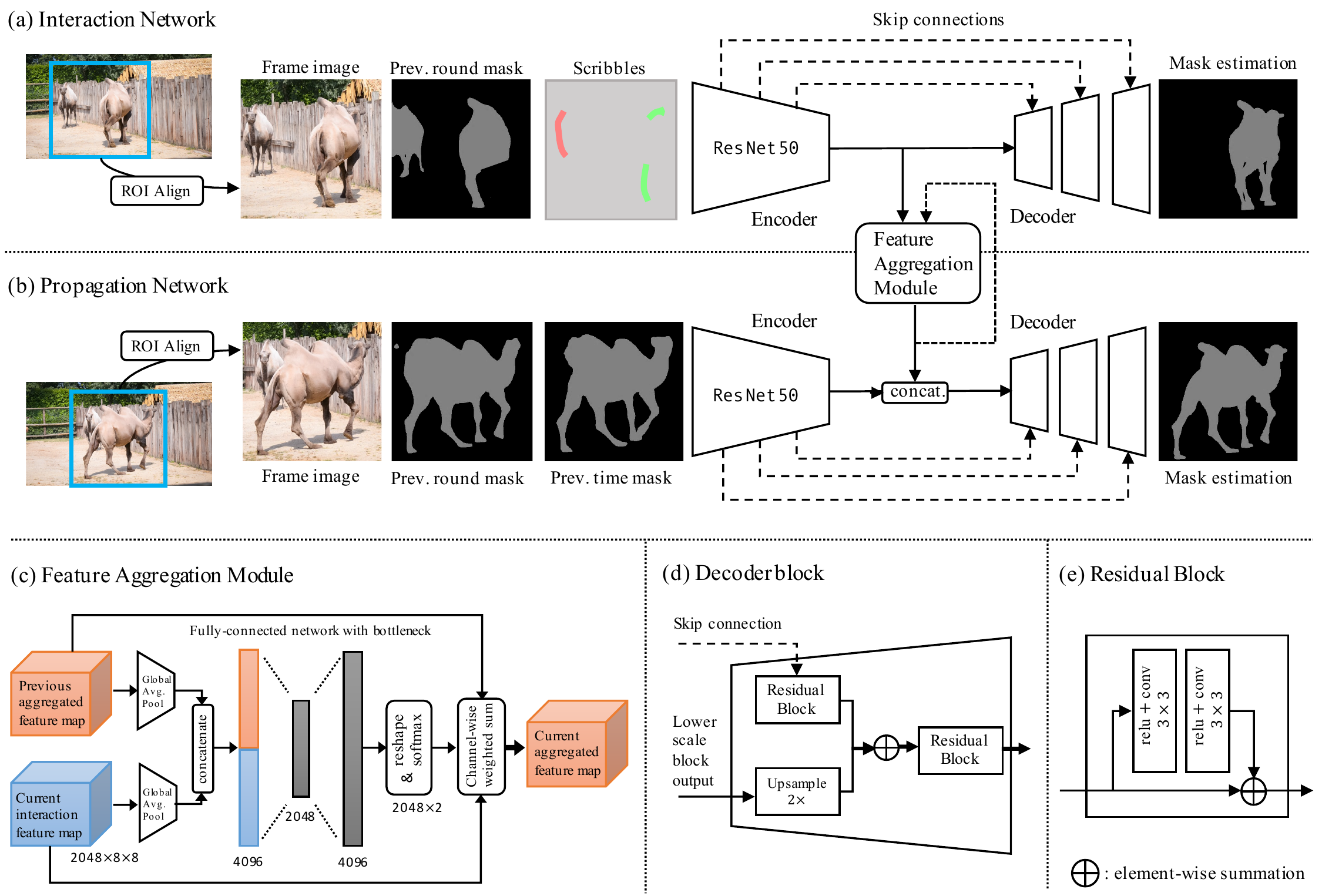}
\caption{The overall network structure. We have two deep networks dedicated each to (a) interaction and (b) propagation tasks. The two networks are internally connected by (c) our feature aggregation module and also externally connected to take the other's output as their input (a, b). Please see \sref{subsec:network} for the details.}
\label{Fig:networks}
\end{figure*}
%%%%%%%%%%%%%%%%%%%%%%%%%

%-------------------------------------------------------------------------
\section{Related Work}
\subsection{Video Object Segmentation} We categorize the video object segmentation into three categories based on different types of user interactions.

\paragraph{Unsupervised Methods.} In the unsupervised setting, there is no user interaction. The unsupervised approaches run automatically but they can only segment visually salient objects based on the appearance or the motion. For example, Jain~\etal\cite{jain2017fusionseg} combine an appearance model with an optical flow model to segment generic objects in videos. Similarly, Tokmakov~\etal\cite{tokmakov2017learning} use a motion estimation network with a recurrent neural network to segment moving foregrounds. The fundamental limitation of the unsupervised methods is that users have no means to select the object of interest.

\paragraph{Semi-supervised Methods.} In the semi-supervised setting, the ground-truth mask of an object in the first frame is provided. The goal is to propagate the object mask throughout the entire video sequence. Many recent approaches~\cite{caelles2017one,voigtlaender2017online,maninis2017video} employ the online learning by fine-tuning deep network models at the testing time in order to remember the appearance of the target object on the given object mask.
Then the object segmentation is performed for each frame. 
%This type of learning is called one-shot learning or online learning. 
Instead of employing the online learning, Jampani~\etal\cite{jampani2016video} propagate the object mask by bilateral filtering. Oh~\etal\cite{oh2018fast} use Siamese two-stream networks and leverage synthetic training data.
Although the semi-supervised methods do not have the limitation of the unsupervised methods, they require a fully annotated object mask in the initial frame, which can be expensive to acquire. 
Additionally, semi-supervised methods rely on extra information such as fully annotated masks or external tools to further improve the output quality.

\paragraph{Interactive Methods.} In the interactive setting, users can provide various types of inputs (\eg bounding box, scribbles, or masks) to select an object of interest in the beginning. 
Users can also provide more interactions to refine the segmentation results.
The goal of this interactive approach is to achieve satisfactory segmentation results with a minimum number of user interactions. 
Many interactive methods~\cite{wang2005interactive, price2009livecut, fan2015jumpcut, bai2009video, bratt2012rotoscoping, li2016roto++} have been proposed. 
\cite{wang2005interactive,price2009livecut,shankar2015video} solve spatio-temporal graphs with hand-crafted energy terms. 
Some methods find the corresponding patches between a target frame and a reference frame, then utilize local classifiers~\cite{bai2009video,zhong2012discontinuity} or an existing patch-match algorithm~\cite{fan2015jumpcut}. \cite{agarwala2004keyframe,li2016roto++} solve the segmentation task by tracking. 
Recently, \cite{benard2017interactive,caelles20182018} proposed deep-learning based methods by modifying semi-supervised methods to the interactive scenario. 
% added to avoid the fatal latex compile error.
Benard and Gygli~\cite{benard2017interactive} use the deep interactive image segmentation method~\cite{xu2016deep} to select an object given initial strokes or clicks, and use the semi-supervised video object segmentation method~\cite{caelles2017one} to propagate the object mask.
Compared to such a simple combination of two separate methods, we carefully design two module networks to interact with each other and train the whole networks jointly using our new multi-round training scheme.

\subsection{Interaction with Deep Neural Networks}
Recently, several methods have been introduced for integrating user interaction with deep neural networks for various interactive tasks.
Xu~\etal~proposed to transform clicks~\cite{xu2016deep} or bounding boxes~\cite{xu2017deep} into Euclidean distance maps for the interactive image segmentation. Zhang~\etal~\cite{zhang2017real} incorporated a user's color selection for the image colorization. Sangkloy~\etal~\cite{sangkloy2017scribbler} and Isola~\etal\cite{isola2017image} used sketches to help generate realistic natural images. 

Different from the above interactive approaches that only consider an interaction given once onto an image, our model considers multiple user inputs possibly drawn onto different video frames. 
The sequence of multiple user interactions is aggregated by a specially designed recurrent block called the feature aggregation module.
In addition, we use the segmentation results from previous rounds as an additional channel, in order to consider the unique characteristics of the interactive video segmentation.
% Inspired by~\cite{xu2016deep}, our algorithm transforms user interactions (\eg scribbles) into additional input channels. 

\section{Method}
Given user annotations on a video frame (\eg scribbles drawn on the foreground and background pixels of an image), we aim for cutting out the target object in all frames of the given video.
From the initial user input, we generate object masks of all frames solely based on the user annotation.
If the user provides additional feedback annotations after reviewing the generated masks, our method refines the object masks based on both additional user annotations and the previous mask estimation results.

To this end, we define two basic operations for the task: interaction and propagation. Two deep CNNs dedicated for each operation are proposed as shown in~\fref{Fig:networks}~(a),(b). 
The interaction network generates the object mask (or refines the previous results) for the annotated frame according to the user inputs. 
The propagation network generates the object masks (or refines the previous results) by temporally propagating the object mask information both forward and backward starting from the frame with user annotation. 

To prevent the error accumulation due to drifts and occlusions during the propagation, the propagation network refers to a reliable visual memory similar to \cite{oh2018fast, yang2018efficient, yoon2017pixel}. While \cite{oh2018fast, yoon2017pixel} employ a Siamese network to access the reference frame directly, we modified the framework to make it more suitable for the interactive video object segmentation.
Specifically, as the most reliable information is contained in the user annotated frames in the interactive scenario, we allow the propagation network to access the features of the interaction network. In addition, we propose a feature aggregation module that accumulates all the previous reference information encoded by the interaction network. This reference-guided propagation is effective, especially for the long-term propagation.

We refer to the series of operations consisting of both the user interactions on one frame and a number of consecutive propagation towards both ends as a \textit{round} (see \fref{Fig:multi-round}). Users are able to repeat several rounds of interactions to refine the segmentation results until they are satisfied with the results as shown in~\fref{Fig:teaser}. Both networks operate on the results obtained from the previous round. We use the same networks for every round. 

\subsection{Network Design}\label{subsec:network}
We have two networks, interaction and propagation, and both networks are constructed as an encoder-decoder structure that can effectively produce a sharp mask output. %~\cite{Lin_2017_CVPR, pinheiro2016learning}. 
We adopt the ROI align before the encoder to make our networks to pay attention to the region of interest (the area around the target object)~\cite{he2017mask}.
We take ResNet50~\cite{he2016deep} (without the last global pooling and fully-connected layers) as the encoder network, and also modify it to be able to take additional input channels (\eg scribbles and the previous masks) by implanting additional filters at the first convolution layer~\cite{perazzi2017learning, xu2016deep}.
The network weights are initialized from the ImageNet pre-trained model, except for the newly added filters which are initialized randomly. 

The decoder takes the output of the encoder and produces an object mask. 
To reconstruct a sharp mask by fully exploiting the information at different scales, the decoder additionally takes intermediate feature maps inside the encoder through skip connections. 
We make modifications to the feature pyramid networks~\cite{Lin_2017_CVPR, pinheiro2016learning} by adding residual blocks~\cite{he2016identity} and use it as the building block of our decoder, as shown in~\fref{Fig:networks} (d),(e). 
The decoder estimates the object mask in a quarter scale of an input image.
For the multi-object scenario where scribbles for each object are given, we first estimate masks for each object then merge the masks to get the multi-object mask using the soft aggregation proposed in~\cite{oh2018fast}.

\paragraph{Interaction Network.} 
The input to the interaction network consists of a frame, the object mask from the previous round (if available), and two binary user annotation maps for the positive and the negative regions respectively. 
The inputs are concatenated along the channel dimension to form an input tensor $\mathbf{X_i} \in \mathbb{R}^{6\times H\times W}$.
The object mask is represented as a probability map filled with values between 0 and 1. 
If no previous mask is available (\eg at the first round), we feed a neutral mask filled with 0.5 for all pixels.
The output of this network is $\hat{\mathbf{Y_i}} \in \mathbb{R}^{H\times W}$, the probabilities of the target object at every pixel. 

\paragraph{Propagation Network.} 
The input to the propagation network consists of a frame, the object mask obtained at the previous frame, and the object mask obtained at the previous round. 
Similar to the interaction network, the inputs are concatenated along the channel dimension to be a tensor $\mathbf{X_p} \in \mathbb{R}^{5\times H\times W}$. 
The two object masks are represented with probabilities and the neutral mask is used if the mask is not available. 
Different from the interaction network, the decoder of this propagation network additionally takes the reference feature map which is computed by our feature aggregation module. 
The reference feature map and the encoder output of this propagation network are concatenated along the channel dimension and are fed into the decoder.  %The feature aggregation module accumulates information of the target object from all the previous rounds.

\paragraph{Feature Aggregation Module.}
In the interactive video object segmentation, the system often takes multiple user annotations in different frames through multiple rounds.
It is important to exploit all previous user inputs for good performance.
To achieve this, we propose a feature aggregation module which is specially designed for accumulating information of the target object from all user interactions.
We use the encoder output of the interaction network to generate reference feature maps. We update the feature maps recurrently when a new user interaction triggers the interaction network.
We design this module to be able to select memorable features by self-attention.
As shown in~\fref{Fig:networks}~(c), the module first performs a global average pooling on the spatial dimension of the feature maps to obtain compact feature vectors.
The vectors are concatenated and fed into two fully-connected layers with a bottleneck.
The outputs of the layers are two channel-wise weight vectors ($\alpha$ and $\beta$) after reshaping and a softmax.
We place the softmax layer to make sure that $\alpha + \beta = \mathbf{1}$.
The two feature maps are channel-wise weighted by $\alpha$ and $\beta$, then merged by the summation: $\mathbf{A}_r = \alpha\odot\mathbf{A}_{r-1} + \beta\odot\mathbf{R}_{r}$.
$\mathbf{A}_r$ and $\mathbf{A}_{r-1}$ are the aggregated reference feature map at the round $r$ and $r-1$ respectively, and $\mathbf{R}_r$ is the encoder output of the interaction network at the round $r$, and $\odot$ is an element-wise multiplication on the channel dimension.

\paragraph{Region of Interest (ROI).}
While fully convolutional networks for image segmentation~\cite{long2015fully} can handle image inputs in any resolution, the performance heavily relies on the absolute scale of objects. 
For example, small objects are easily missed and objects larger than the receptive field need to be estimated by observing only a part of the objects.
This issue can be addressed when the network knows where to look. In our case, we can reason about the region of interest (ROI) from the guidance (\eg scribbles and masks). 

To take advantage of the guidance, we first compute a tight box that contains all available guiding information (which include user scribbles, the mask from the previous frame, and the mask from the previous round) and set the ROI to a box that is computed by doubling each side of the tight box.
Then, the ROI area for all the inputs is bilinearly warped into a fixed size (\eg $256\times256$ in our implementation) before we feed them into the encoders~\cite{jaderberg2015spatial, he2017mask}.
Finally, the prediction made within the ROI is inversely warped and pasted back to the original location.
The training losses become scale-invariant as they are computed in the ROI-aligned space, and this enables us to not use the complex balanced loss functions~\cite{caelles2017one}. Note that we set ROI as the whole image at the first round and start to compute ROI using the guidance from the second round. 

% \paragraph{Multi-object segmentation.}
% While above description is based on a single object case, we can easily extend our method to handle multi-object segmentation. 
% For the multi-object scenario where scribbles for each object are given, we first estimate masks for each object then merge the masks to get the multi-object mask using the soft aggregation proposed in~\cite{oh2017fast}:
% %
% \begin{multline}
% p_{i,m} = \sigma\big(l(\hat{p}_{i,m})\big)
% = \frac{\hat{p}_{i,m}/(1-\hat{p}_{i,m})}{\sum_{j=0}^{M}\hat{p}_{i,j}/(1-\hat{p}_{i,j})},
% \\
% \text{s.t.}\quad
% \hat{p}_{i,0} = \Pi_{j=1}^M(1-\hat{p}_{i,j}),
% \label{Eq:softmax}
% \end{multline}
% %
% where $\sigma$ and $l$ represent the softmax and logit functions respectively, $\hat{p}_{i,m}$ is the network output probability of the object $m$ at the pixel location $i$, $m$=$0$ indicates the background, and $M$ is the total number of objects.
% When our model infers masks for each object, we provide additional mask information that indicates pixels belonging to other objects. 
% The series of these operations can be differentiably connected, and our model is fine-tuned using videos with multi-instance masks in an end-to-end manner. 

\subsection{Training}

\begin{figure}
\centering
\includegraphics[width=1.0\linewidth]{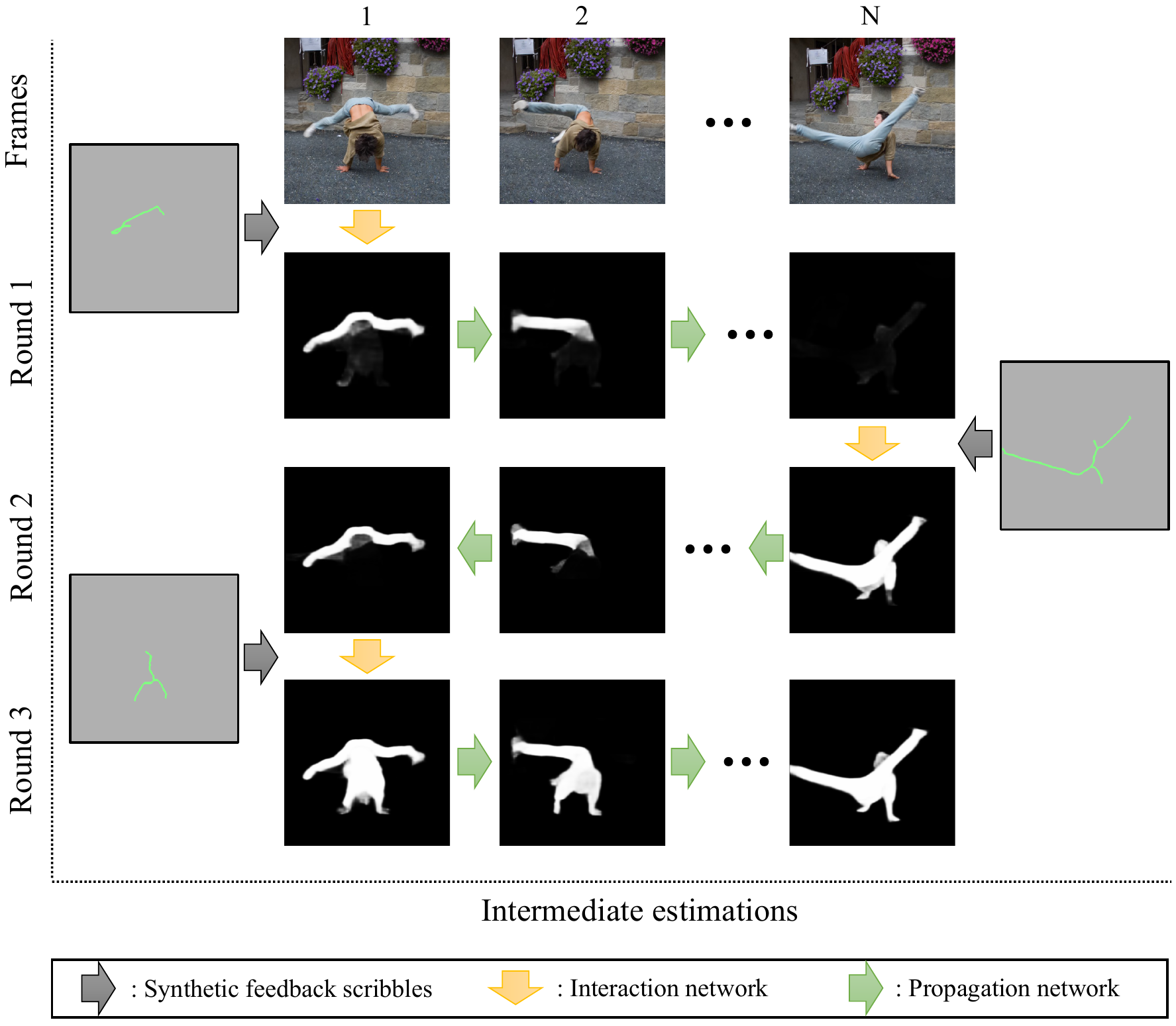}
\caption{An example of a single training iteration in our multi-round training scheme. The multiple rounds of the network feed-forwarding form a single training iteration so that the networks can experience a real testing scenario and learn how to understand user intention and update incorrect estimations. Training losses are computed at every intermediate estimation.}
\label{Fig:multi-round}
\end{figure}

\paragraph{Multi-round Training.} 
For the best testing performance, we make our training loop close to the real testing scenario: a user interacts with our model multiple times while providing feedback in the forms of scribbles on multiple frames. 
We propose a new multi-round training scheme where a single training sample consists of multiple rounds of user interactions. 
At every round, our model is trained to refine the previous round's results by understanding the user's intention (interaction network) and temporally propagating the object mask (propagation network).
Two networks are trained jointly by making an estimation using the previous estimation that can be inferred from the other network. 
Losses are computed at every intermediate prediction and the back-propagation is performed at every loss computation to update the parameters of the networks. 
At each round, user inputs are synthesized by simulating user behaviors. 
\fref{Fig:multi-round} shows an example of a single training iteration in our multi-round training scheme.

%%%%%% figure %%%%%%%%%%%
%%%%%%%%%%%%%%%%%%%%%%%%%

\paragraph{User Scribble Synthesis.} 
One challenge in training an interactive model is collecting user input data. 
For our scenario where a user provides scribbles as feedback, it is not feasible to collect large training data. 
Instead, we train our model with synthetically generated user interactions. 
In the first round, positive scribbles are sampled from the foreground region.
In the following rounds, scribbles are synthesized within false negative and false positive areas where the areas are computed using the ground-truth mask. 
We sample positive scribbles from the false negative area and negative scribbles from the false positive area. 

We use morphological skeletonization to automatically generate realistic scribbles similar to \cite{caelles20182018}.
Given a candidate area to sample scribbles, we first remove small false estimations isolated from the main body by repeating a binary morphological opening operation. 
Then, we perform the skeletonization of the mask to get either positive and negative scribbles within the target area. We use a fast implementation of the thinning algorithm~\cite{Guo:1989:PTT:62065.62074} for the skeletonization.

A concern can be raised about the gap between the simulated and the real scribbles. 
We empirically validate that our model trained with simulated user scribbles works well with real user interactions as shown in our demo video.

\paragraph{Pre-training on Images.}
It is widely known that training deep networks requires a large amount of data. 
However, video data that comes with object masks are limited due to laborious human annotation process. 
We bypass the issue by employing two-stage training where our networks are first pre-trained on synthetic image data and then are fine-tuned on real video data. 
The idea that trains a video segmentation network on image data was proposed in~\cite{perazzi2017learning}, and we follow the data simulation method in~\cite{oh2018fast}. 
The method produces a set of reference and target frame pairs by applying random affine transforms and object composition. 
This pre-training is similar to training on videos, but temporal propagation is limited to a single step as there are no consecutive frames. 

\paragraph{Implementation Details.}
For the pre-training, we combine multiple image datasets that come with object masks (salient object detection -- \cite{shi2016hierarchical, cheng2015global}, semantic segmentation -- \cite{everingham2010pascal, hariharan2011semantic, lin2014microsoft}). After the pre-training, we use the video data from the training subset of DAVIS~\cite{Pont-Tuset_arXiv_2017}, GyGo~\cite{GyGo}, and Youtube-VOS~\cite{xu2018youtube} to train our networks. 

To sample training data, we first resize video frames to be 480-pixels on the shorter edge while keeping the aspect ratio. 
Then, $N$ consecutive 400 $\times$ 400 sized patches are sampled from a random location of the video, where $N$ is the length of a training video clip. 
We randomly skipped frames to simulate fast motion and $N$ is gradually increased from 4 to 8 during training. 
We also augment all the training samples using random affine transforms. 
The number of rounds also grows from 1 to 3 during training. 
The loss is computed by the cross-entropy function and we use Adam optimizer with a fixed learning rate of 1e-5. 
The training with video data takes about 5 days using a single NVIDIA GeForce 1080 Ti GPU.

%%%%%% figure %%%%%%%%%%%
\begin{figure}
\centering
\includegraphics[width=\linewidth]{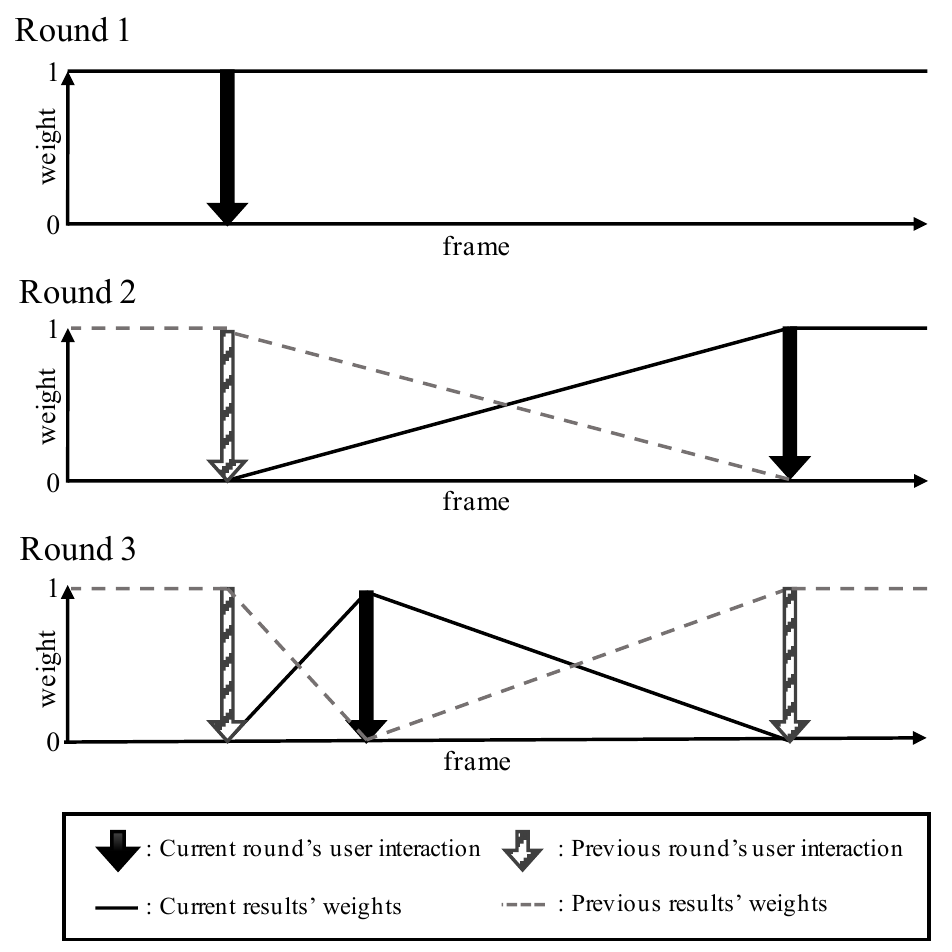}
\caption{Round-based Testing Scheme. At each round, we update previous object masks with new estimations by the weighted averaging. Solid lines and dashed lines indicate mask updating weights for current estimation and previous estimation, respectively. Weights are inversely proportional to the propagated distance.}
\label{Fig:testing}
\end{figure}
%%%%%%%%%%%%%%%%%%%%%%%%%

\subsection{Testing Scheme}
%As depicted in~\fref{Fig:teaser}, we adopt the \textit{round-based} interaction scenario proposed in~\cite{caelles20182018}. Our networks perform the two core %operations (interaction and propagation) of the scenario. At the beginning of each round, a user selects a frame and draws scribbles on it. The interaction network %produces an object mask for the selected frame. Then, the propagation network successively propagates the mask both forward and backward until it reaches the end. A %number of rounds can be repeated until the user stops providing additional scribbles. 

One potential issue observed during our testing is that the propagated mask may be worse than the mask from the previous round. This happens especially when the destination is far from the user-selected frame.% where new user annotations are drawn. 
We conjecture that the long-term propagation may be unstable as our model is trained on short video clips. To address this issue, we modified our testing scheme in two ways; continuous updating and restricted propagation. In continuous updating, we update the previous round's masks with newly estimated masks by the weighted average. The weighting factor is inversely proportional to the propagated distance, and different weighting functions such as a linear and the Gaussian were tested. We empirically found that the different weighting functions end up giving similar performance. We used a simple linear function in our experiments. For the restricted propagation, we propagate the object mask until we reach a frame in which user annotations were given in any previous rounds. The restricted propagation improves not only the accuracy by preventing the drift, but also the runtime speed since it requires a smaller number of propagations. This testing scheme is depicted in~\fref{Fig:testing}. 

%Please watch our supplementary video that records our demos using our prototype application for each testing scenario.

\section{Experiments}
%\subsection{Quantitative Evaluation}
It is difficult to evaluate interactive video object segmentation methods quantitatively because the user input is directly related to the segmentation results, and vice versa. 
%For example, object masks are estimated with a series of user annotations and a user can provide additional annotations based on the current masks. 
To tackle this problem with the evaluation, Caelles~\etal\cite{caelles20182018} introduced a robot agent service that simulates human interaction according to the intermediate results of an algorithm. We used their method to quantitatively evaluate our method. 

% \paragraph{DAVIS Challenge.}
\subsection{DAVIS Challenge}
To fairly compare our method against the state-of-the-art methods, we evaluated our model on the interactive track benchmark in the DAVIS Challenge 2018~\cite{caelles20182018}. In the challenge, each method can interact with a robot agent up to 8 times and is expected to compute masks within 30 seconds per object for each interaction. The performance of each method is evaluated using two metrics: area under the curve (AUC) and Jaccard at 60 seconds (J@60s). AUC is designed to measure the overall accuracy of the evaluation. J@60 measures the accuracy with a limited time budget (60 seconds). We summarize the evaluation results in \Tref{Table:challenge}. In both metrics, our method outperforms competing methods by a large margin~\cite{DAVIS2018-Interactive-1st}.

%%%%%% figure %%%%%%%%%%%
\begin{table}
\centering
\begin{tabular}{l?{1pt} c c  } 
% \specialrule{.12em}{.01em}{.01em} 
% \hline
Method &  AUC & J@60   \\ 
\specialrule{.12em}{.01em}{.01em} 
Ours & \textbf{0.641} & \textbf{0.647} \\
Najafi~\etal~\cite{DAVIS2018-Interactive-2nd}  & 0.549 & 0.395 \\
Lin~\etal & 0.450 & 0.240 \\
Huang~\etal & 0.328 & 0.335 \\
Scribble-OSVOS~\cite{caelles20182018} & 0.299 & 0.153 \\
Rakelly~\etal & 0.269 & 0.273 \\
% \specialrule{.12em}{.01em}{.01em} 
% \hline
\end{tabular}
\caption{The leaderboard of the interactive track in the DAVIS challenge 2018. The entries are ordered according to the AUC score. Scribble-OSVOS is a baseline method proposed by the challenge organizer~\cite{caelles20182018}.}
\label{Table:challenge}
\end{table}
%%%%%%%%%%%%%%%%%%%%%%%%%
%%%%%% figure %%%%%%%%%%%
\begin{figure*}
\centering
\vspace{0.3cm}
\includegraphics[width=1.0\linewidth]{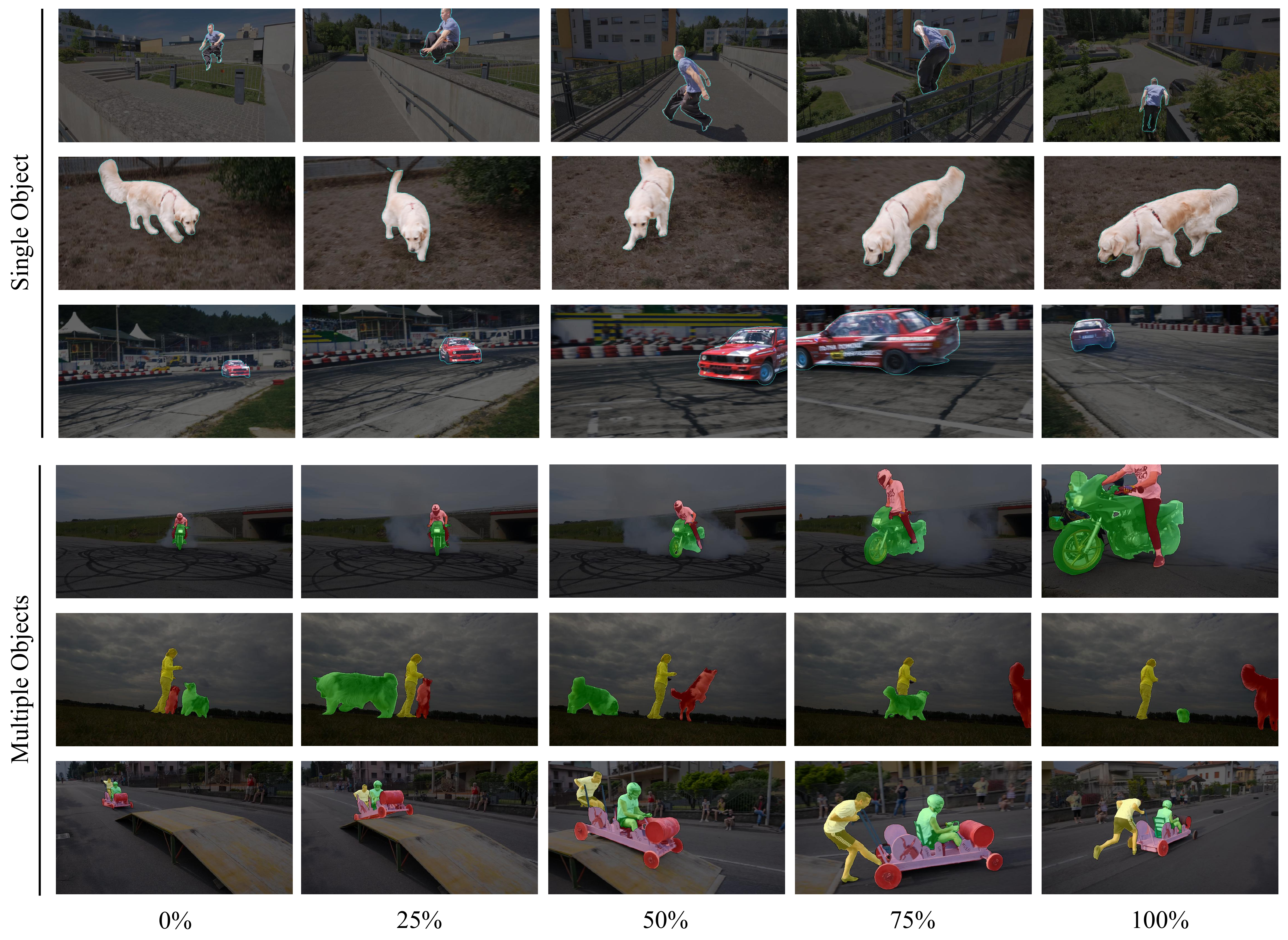}
\caption{The qualitative results on the DAVIS-2017 validation set. All the user interactions are automatically simulated by the robot agent provided by~\cite{caelles20182018}. The result masks are overlaid to uniformly sampled frames after 5 interactions (rounds).}
\vspace{0.1cm}
\label{Fig:qualitative}
\end{figure*}
%%%%%%%%%%%%%%%%%%%%%%%%%
% \paragraph{Qualitative Results.}
\subsection{Qualitative Results}
%\paragraph{Visual Results. }
\fref{Fig:qualitative} shows examples of our results obtained after 5 interactions with the automatic evaluation robot in the DAVIS Challenge 2018. Our method generates accurate segmentation results for various object types with complex motions even if there are multiple object instances. 
In the supplementary video, we present the recording of our real-time demo with real user interactions.  

% \paragraph{Ablation Study.}
\subsection{Ablation Studies}
We conduct an ablation studies using the DAVIS-2017 validation set to validate the effectiveness of our feature aggregation module and training scheme. Specifically, we compare our complete model with three variant models. \textit{No Reference} is a model without the feature aggregation module. In \textit{No Aggregation} model, the feature aggregation module is replaced with a simple identity connection without feature aggregation. \textit{No Multi-Round} is a model trained with the number of rounds as one (\ie at each training iteration, there is only one interaction from the user).

\begin{figure}
\centering
\includegraphics[width=1.0\linewidth]{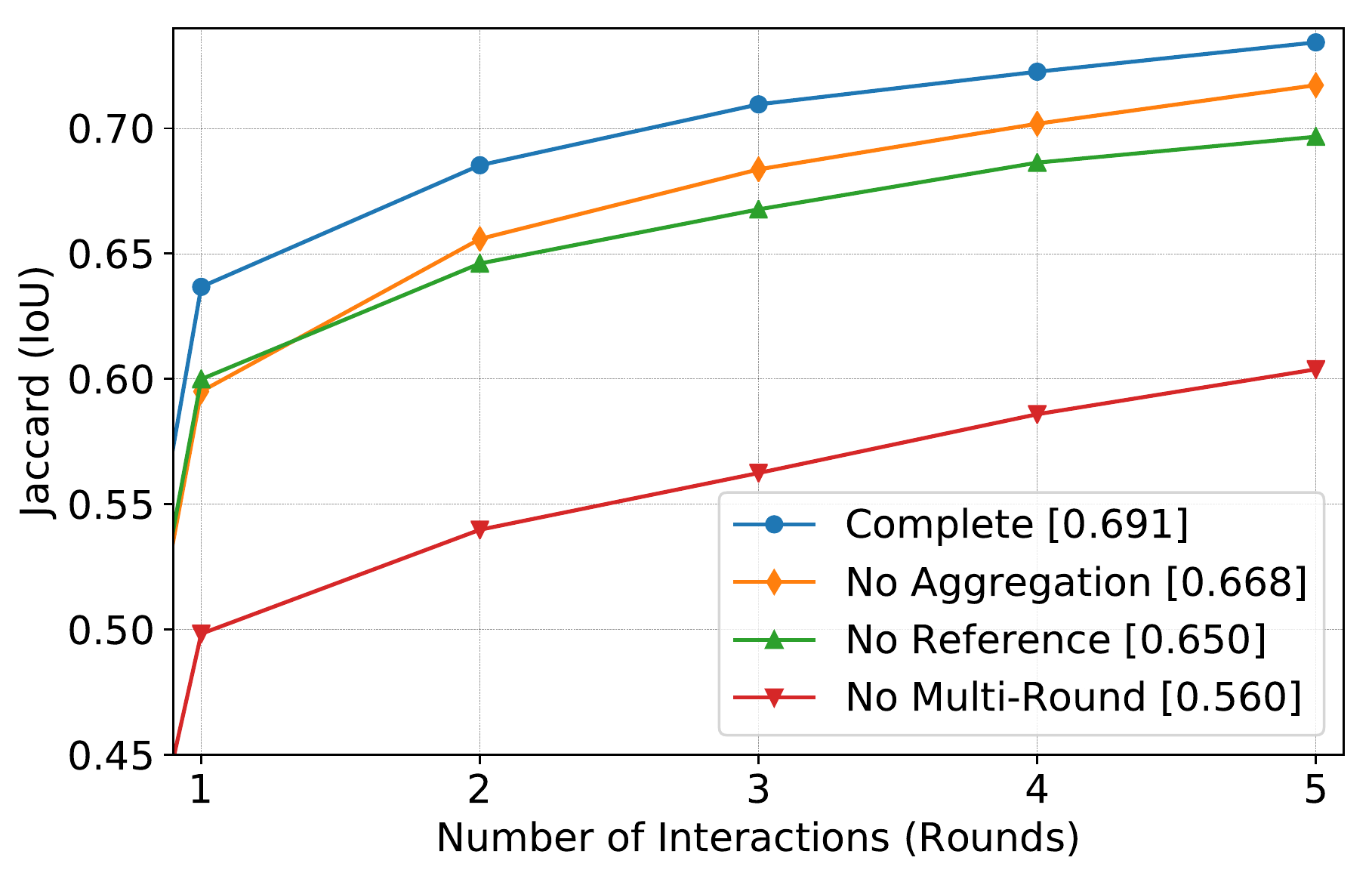}
\caption{The result of our ablation study on the DAVIS-2017 validation set. We compare models with ablations from our complete model. The AUC of each variant is shown in the squared brackets of the legend.}
\label{Fig:ablation}
\end{figure}

%The propagation network takes the reference information which is computed inside the interaction network. 
%Feature maps from the interaction network first go through our proposed feature aggregation module to form an accumulated knowledge through multiple rounds rather than being used directly. 
%To reveal the effect of this mechanism, we conducted ablation studies with different network structures and multi-round training strategy.
%Specifically, we first demonstrate the effect of the reference guide by cutting out the reference feature maps provided to the propagation network. 
%We named this model \textit{No Reference}. 
%Then, we further test the effect of feature aggregation by providing the propagation network with only the recent reference feature maps without feature aggregation. This variant is named as \textit{No Aggregation}. 
%Finally, we tested how our proposed multi-round training influences the results. 
%Model \textit{No Multi-Round} is trained with the number of rounds as one (\ie at each training iteration, there is only single interaction from the user).
The Jaccard score of ablation models with growing number of interactions is shown in~\fref{Fig:ablation}. 
As shown in~\fref{Fig:ablation}, the proposed multi-round training is crucial for achieving high accuracy and our feature aggregation module further improves the performance by allowing the networks to exploit the reference information from all previous user inputs.

Another ablation study was conducted on the use of the training data.
Our complete model is first pre-trained on static image data and then fine-tuned using video data. 
To validate the effect of the pre-training, we compare variant models that are just trained on the video data without the pre-training. Also, to further inspect the effect of the amount of video training data, we evaluate variants that are fine-tuned with only 60 train videos of DAVIS-2017. \Tref{Table:training} summarizes the results obtained by our variant models trained using different combinations of training datasets. Without pre-training, our performance drops significantly. The use of additional training video data further raises our performance.

%%%%%% figure %%%%%%%%%%%
%%%%%%%%%%%%%%%%%%%%%%%%%

\begin{table}
% \small
% \setlength{\tabcolsep}{4pt}
% \renewcommand{\arraystretch}{1.0}
\centering 
\begin{tabular}{ccc|cc}
%\toprule
%\multicolumn{3}{c|}{Training data} & \multicolumn{2}{c}{Measure} \\
%\cmidrule(lr){1-3}
%\cmidrule(lr){4-5}
 PT & DV & GG+YV & AUC  & J@60s \\
\specialrule{.12em}{.01em}{.01em} 
%\midrule
  & \checkmark &  & 0.555 & 0.589 \\
 \checkmark & \checkmark &  &  0.621 & 0.637 \\
  & \checkmark & \checkmark &  0.627 & 0.657 \\
 \checkmark & \checkmark & \checkmark &  \textbf{0.691} & \textbf{0.734} \\
% % Our &   & \textbf{89.0} & \textbf{89.8} & 0.0$s$ \\ % with cc
%\bottomrule
\end{tabular}
\caption{We compare our models trained with different combination of training datasets. PT: pre-training on static images~\cite{shi2016hierarchical, cheng2015global,everingham2010pascal, hariharan2011semantic, lin2014microsoft}. DV, GG and YV: the use of DAVIS~\cite{Pont-Tuset_arXiv_2017}, GyGo~\cite{GyGo}, and Youtube-VOS~\cite{xu2018youtube} for fine-tuning.}
\label{Table:training}
\end{table}
%%%%%%%%%%%%%%%%%%%%%%%%%

% \paragraph{Failure Cases.}
\subsection{Failure Cases}
\begin{figure}
\centering
\includegraphics[width=1.0\linewidth]{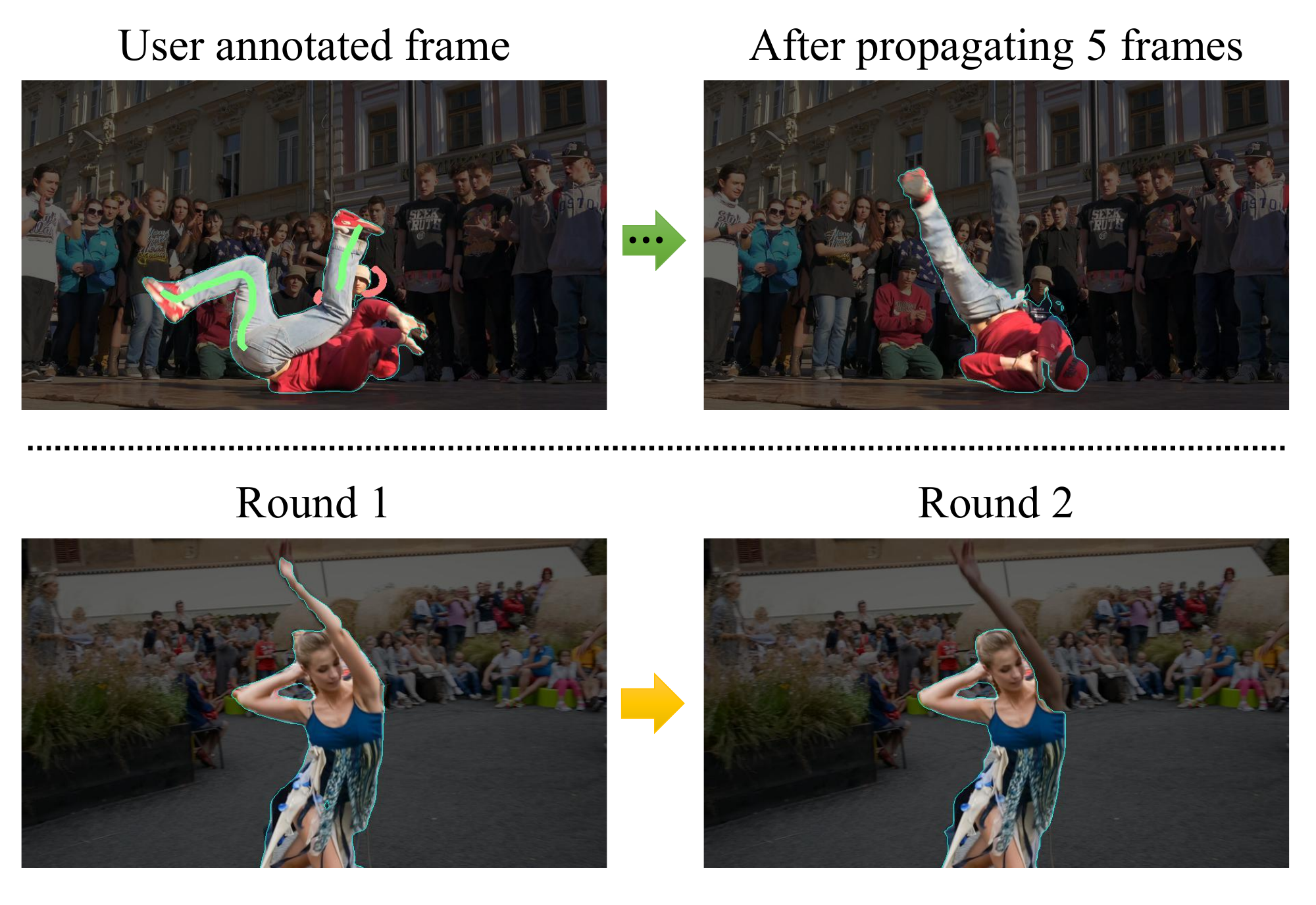}
% \vspace{-5pt}
\caption{Failure cases. (top) Our propagation network may suffer from error accumulation due to fast and complex object motions. (bottom) We take only partial annotations from users and misunderstanding of the user intention may lead to unstable prediction with additional annotations.}
\vspace{-5pt}
\label{Fig:failure}
\end{figure}

While our method demonstrates satisfactory results on both the quantitative and the qualitative evaluations, we found few failure cases as shown in~\fref{Fig:failure}. We observed that rapid and complex object motions may lead our propagation network to drift by the error accumulating as shown in \fref{Fig:failure}~(top). We believe that a good future direction is to augment the algorithm with a reliable temporal propagation of object masks.

Another limitation we found is that our method may be less stable on very challenging scenes in the current round-based scenario. Our method mostly improves results with additional user interactions, but this is not guaranteed as shown in~\fref{Fig:failure}~(bottom).
Since we take only partial annotations from users at each round, the propagated masks from newer round are sometimes less accurate and there is no guarantee that we can always keep better results from different rounds.
% Since we take only partial annotations from users at each round, and misunderstanding of the user intention may lead to worse results with additional annotations.
This is because there is no safety gear in the testing scenario and it can be resolved by asking the user for the confirmation of the mask being good to prevent updating the masks. 

% Ning: the propagated features from newer round sometimes provide less useful information and there is no guarantee that we can always keep better results from different rounds. This suggests that we can improve the performance by either asking users to verify the results as in the rotoscroping scenario or learning to combine the propagated information instead of the linear combination strategy, which we leave for future study.

%%%%%% figure %%%%%%%%%%%
%%%%%%%%%%%%%%%%%%%%%%%%%

\section{Conclusion}
While object segmentation in a video is one of the most basic tasks for video editing, it requires a lot of user effort and time with existing tools. To make it more accessible, we have presented a novel technique that generates object segmentation masks in video frames with minimum user inputs. Our method consists of interaction and propagation networks that share information with the feature aggregation module. We proposed the multi-round training scheme designed for interactive tasks and it plays a key role in achieving high accuracy. While our model is trained using synthetic user interactions, our method not only shows the best performance on the quantitative evaluation but also demonstrates good performance with real user interactions. %We demonstrate our system is flexible to adopt two different style of user interaction.

There are directions to further improve our system. The drifting during propagation is still a major challenge, although we greatly improved the performance with the aggregated reference features and the multi-round training. We believe that a better semantic understanding of the scene will help to resolve this problem by robustly linking the instances with appearance changes across video frames. Another important future work is supporting high-resolution videos. This is one of the common issues in many deep learning-based segmentation algorithms, and we hope that this can be addressed with a better network architecture or by combining our work with additional post-processing modules.\\

% \paragraph{Acknowledgement}\\

\noindent\textbf{\large{Acknowledgement}}\vspace{3pt}\\
This work was supported by Institute for Information \& communications Technology Promotion (IITP) grant funded by the Korea government (MSIP) (2018-0-01858).
% , Video Manipulation and Language-based Image Editing Technique for Detecting Manipulated Image/Video).

{\small
\bibliographystyle{ieee}
\bibliography{egbib}
}

\end{document}